\documentclass[11pt]{article}

\usepackage[final]{acl}

\usepackage{times}
\usepackage{latexsym}

\usepackage[T1]{fontenc}
\usepackage[utf8]{inputenc}

\usepackage{microtype}

\usepackage{inconsolata}

\usepackage{graphicx}

\usepackage{booktabs}
\usepackage{array}
\usepackage[misc]{ifsym}
\usepackage{CJKutf8}

\usepackage{subcaption}
\usepackage{amsmath}
\usepackage{float}
\usepackage{fontawesome5}

\title{Beyond Surface Cues: \\Disentangling Sociocultural Signals in Multilingual LLMs}

\author{
    Yuanjun Feng\textsuperscript{\rm 1},
    Tanzhou Liu\textsuperscript{\rm 1},
    Stefan Feuerriegel\textsuperscript{\rm 2},
    Yash Raj Shrestha\textsuperscript{\rm 1} \\
    \textsuperscript{\rm 1}University of Lausanne, Switzerland \\
    \textsuperscript{\rm 2}LMU Munich, Munich Center for Machine Learning (MCML), Germany \\
    \texttt{\{yuanjun.feng, tanzhou.liu, yashraj.shrestha\}@unil.ch, feuerriegel@lmu.de}
}

\begin{document}
\maketitle

\begin{abstract}
Multilingual LLM outputs can vary across sociocultural contexts. However, evidence of cultural grounding can be misleading: identity labels may be inferred from explicit or indirect textual cues, while names and wording can reveal the source language. Treating all these signals as evidence of cultural grounding may obscure potential biases. We present a human-validated, multi-agent audit that separates three questions: whether outputs reproduce social biases, whether identity groups are represented differently, and whether outputs reflect cross-cultural patterns. The study analyzes 89,253 outputs from 12 LLMs in English, French, and Chinese, spanning 18 occupations and three task conditions.

We find that bias representation varies systematically across languages and tasks. Removing direct identity cues sharply reduces identity-label prediction in English and Chinese, but has a much smaller effect in French. Across all language–genre settings, the cultural context associated with the source language receives the highest average relevance score, with moderate agreement between automated and human ratings. However, the ability to identify the source language drops substantially after translation and again after masking names. Without these controls, multilingual audits may mistake surface cues for cultural understanding, leading to misleading conclusions about cross-cultural variation and bias. Our audit offers a practical framework for separating such shortcuts from more meaningful cross-cultural patterns.

\end{abstract}

\section{Introduction}

In the real world, bias rarely announces itself with a single, explicit sentence. Instead, it is deeply embedded within the complex fabric of narratives, manifesting in the allocation of roles, the logic of causality, and the subtlety of value judgments \cite{tabassum2021gender,santoniccolo2023gender}. This invisible current continuously shapes readers' commonsense assumptions about who seems naturally suited to which occupations. Rather than overt declarations of gender superiority, bias operates through the relentless repetition of assumptions about who is portrayed as a rational protagonist versus who is relegated to supportive, emotional labour \cite{allan_stereotypical_2025,rettberg2025ai}. Such inconspicuous yet cumulative narrative bias is difficult to detect and correct in digital media environments, and the assumptions it conveys can take hold early and shape subsequent interests \cite{bian2017gender}.

As large language models (LLMs) increasingly serve as the narrative infrastructure of human communication, they actively shape occupational and cultural narratives. In these settings, models do not merely ``retrieve facts''; they generate explanations that carry culturally specific assumptions across languages. These narratives can reinforce existing inequalities and risk propagating dominant cultural norms when producing content for multilingual audiences \cite{kirk2021bias}.

Many studies demonstrate that LLMs inherit and often amplify sociocultural bias in their training data \cite{bolukbasi_man_2016,zhao_gender_2018}. However, existing evaluations predominantly focus on explicit bias, assessed through sentence-completion prompts or sentence-level completions \cite{kotek_gender_2023}. This focus on explicit bias, though useful, overlooks implicit and hidden biases that arise in complex contexts such as task allocations, persona settings, and nuanced language choices \cite{wilson_gender_2024}. Identifying these subtler forms of bias is essential to developing fairer and more reliable language models. 

Moreover, a fundamental tension persists between the pursuit of universal fairness and robustness and the preservation of culturally grounded diversity. Although recent discourse has shifted from identifying direct harms to examining how LLMs shape social perceptions \cite{weidinger2021ethical,lin2025implicit}, their cultural impact remains insufficiently understood. In particular, it is unclear whether biases measured in explicit, controlled settings persist in implicit, long-form contexts, and how these dynamics vary across languages that encode distinct discursive norms and cultural traditions \cite{ghosh2025bias}. As LLMs are deployed globally, there is still limited systematic evidence on how sociocultural patterns in generated narratives vary across linguistic settings and whether apparent cross-cultural differences persist after surface cues are controlled.

To bridge this gap in linguistic and cultural understanding, we introduce a multi-agent framework to audit sociocultural patterns in LLM-generated narratives. Our framework connects three complementary analytical lenses: bias representation, identity-linked semantic separation, and cross-cultural patterns.

We organize the audit around these lenses:
\begin{center}
\fbox{%
 \parbox{0.96\linewidth}{%
 \small
 \setlength{\parskip}{0.3em}
 \textbf{Bias Representation:} examines how socially patterned associations vary across language-linked contexts and task conditions.

 \textbf{Identity-Linked Semantic Separation:} uses full-text label recovery as a positive control and examines how identity-label recoverability changes after direct-cue deletion across languages.

 \textbf{Cross-Cultural Patterns:} examines context-specific relevance alongside the sensitivity of source-language separability to surface-cue controls.
 }%
}
\end{center}

\begin{figure*}[htbp]
    \centering
    \includegraphics[width=0.95\textwidth]{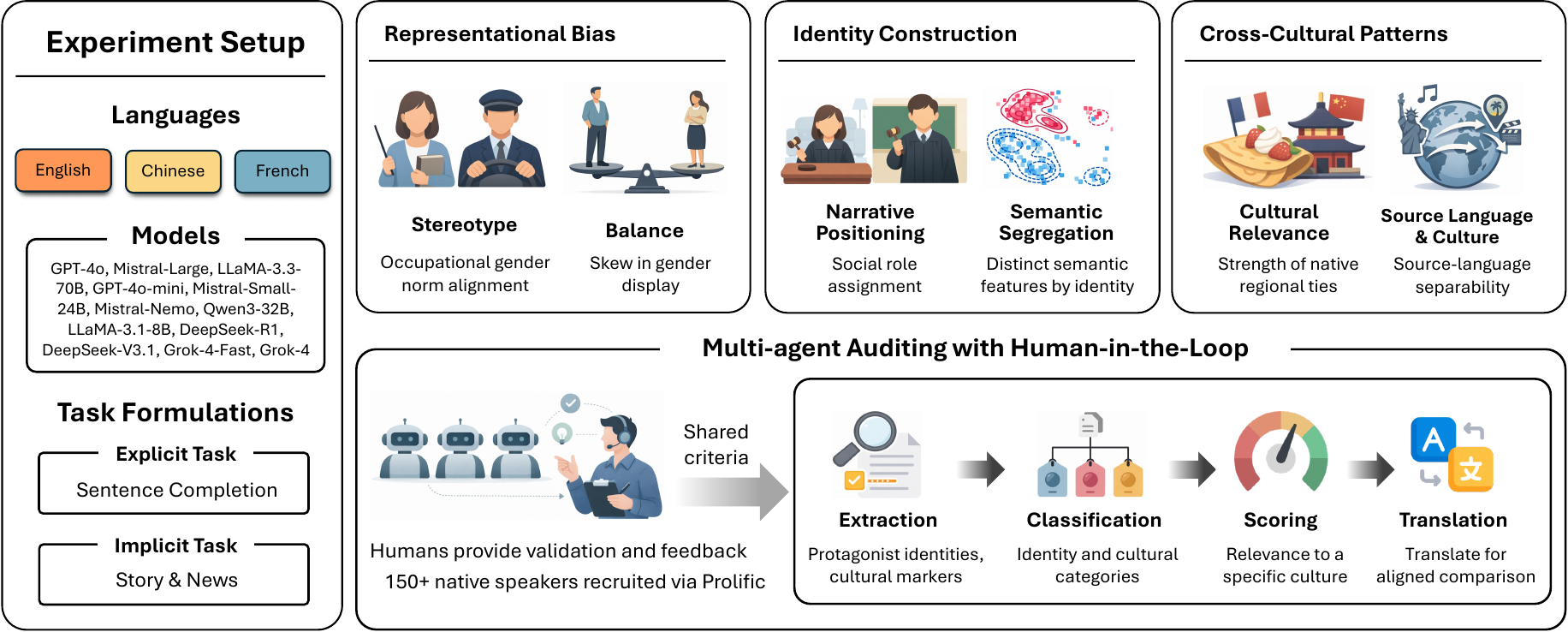}
    \caption{Overview of the experiment setup.}
    \label{fig:design}
\end{figure*}

Figure~\ref{fig:design} provides an overview of the experiment setup and multi-agent auditing workflow.

Appendix~\ref{app:lens-evidence-map} maps each analytical lens to its study-specific focus and evidence.

\textbf{Contributions:}
(1) \textbf{Multi-level sociocultural framework:} We conceptualize multilingual sociocultural evaluation as a multi-level problem spanning bias representation, identity-linked semantic separation, and cross-cultural patterns, and operationalize these levels in a unified audit.
(2) \textbf{Empirical findings:} Our experiments show that language and task shape how LLMs represent social bias, identity, and cultural context, while cue-removal controls show that some apparent differences rely on names and language-specific wording; cultural grounding should therefore be evaluated through meaning and contextual relevance rather than surface cues alone.

\section{Related Work}

As LLMs increasingly function as the narrative infrastructure of human communication, their reliance on vast, uncurated textual data exposes them to deeply embedded gender and occupational stereotypes \cite{bolukbasi_man_2016,caliskan_semantics_2017}. This raises significant concerns regarding representational bias, particularly in socially impactful contexts such as career-path recommendations and résumé screening \cite{wilson_gender_2024,gaebler_auditing_2024}, and in the political framing of automatically generated content \cite{bang_measuring_2024,motoki_more_2024}.

Existing research on LLM bias can be divided by task formulation. One stream uses controlled, explicit prompts and sentence-completion benchmarks to detect isolated bias patterns \cite{nangia_crows-pairs_2020,nadeem_stereoset_2021}. \emph{WinoBias}, for example, tests associations between occupations and gendered pronouns in sentence completions \cite{zhao_gender_2018,zhao-etal-2019-gender}. However, these isolated metrics often miss how bias surfaces in long-form generation. Another stream focuses on implicit tasks. Here, bias is not overt but woven into the logic and evaluative framing of stories or news reports \cite{hofmann_ai_2024,rettberg2025ai}. In these settings, prompts can shape the semantic construction of generated identities \cite{steinborn_information-theoretic_2022,gnadt_exploring_2025}.

The influence of language structure on the reproduction of gender stereotypes is deeply connected to cultural context and the composition of training data \cite{abid_persistent_2021,zhong_cultural_2024}. Language and culture intertwine, as cultural dimensions like gender norms shape institutions and civic participation \cite{alesina_origins_2013}. Grammatical gender is a key linguistic driver. In gendered languages such as French, gender is morphologically expressed in nouns and adjectives. Non-gendered languages (such as English) or character-based languages (such as Chinese) employ different morphological or semantic strategies. Cross-country evidence suggests that gendered language structure correlates with higher societal gender inequality \cite{mavisakalyan_gender_2015}. Evidence from immigrant households further suggests that speakers of more gender-marked mother tongues divide household labour along more traditional lines \cite{hicks_does_2015}.

LLM development, especially Reinforcement Learning from Human Feedback (RLHF), adds complexity. RLHF can suppress overtly biased language while complying with safety policies \cite{dai_safe_2023}. However, RLHF yields only limited improvements on bias benchmarks \cite{ouyang2022training}, and reduced overt bias need not imply the absence of gender-associated differences in longer-form content. This motivates measuring both aggregate representation and separation in narrative embedding space.

To evaluate and mitigate these biases, studies have shifted from static evaluations \cite{kurita_measuring_2019,cryan_detecting_2020} toward dynamic and agentic approaches. Tools such as \emph{RUTEd} \cite{lum_bias_2025}, multi-agent debate frameworks \cite{feng-etal-2025-mad}, and simulations of interaction with stereotypically biased AI \cite{allan_stereotypical_2025} provide infrastructure for diagnosing long-form behavior. Yet multilingual cultural audits introduce an additional identification problem. A marker may reveal its source language through original-language wording, names, institutions, or places; treating a language as a culture can also erase within-language heterogeneity \cite{hershcovich-etal-2022-challenges}. Existing work documents cross-cultural preferences and cultural dominance, but rarely tests source-language separability under translation and named-entity masking. Our analysis pairs three-context relevance scores with those two controls.

\section{Methodology}

We conduct a multilingual evaluation of sociocultural patterns in LLM-generated outputs under different task conditions. The audit is organized around three complementary analytical lenses---bias representation, identity-linked semantic separation, and cross-cultural patterns.

\subsection{Multi-Agent Auditing Framework}

The framework coordinates four agents: extraction, classification, cultural scoring, and translation. They identify protagonist labels and text-grounded cultural markers, select and score markers in three contexts, and prepare non-English markers for the surface-cue controls. Figure~\ref{fig:design} summarizes output generation, framework processing, and the subsequent three-lens analysis; full agent roles and scoring settings appear in Appendix~\ref{app:method-details}.

Two Prolific human studies validate cultural-marker annotations (39 raters; 624 marker-level comparisons) and protagonist-label extraction (116 raters). Human endorsement of automated marker-inclusion decisions is 80.0\%. After weighting to match the full-corpus distribution, automated and human protagonist labels agree for 91.3\% of EN, 82.6\% of FR, and 81.6\% of ZH cases. Appendix~\ref{app:method-details} details sampling, aggregation, agreement metrics, and resampling procedures.

\subsection{Materials}
\label{sec:materials} 

\subsubsection{Task Formulations}
 We use occupations as a focused and socially consequential case domain for measuring representational bias. Diverse professions provide semantic variation and are intertwined with sociocultural factors such as gender norms, power, and stereotypes \cite{kirk2021bias}.

 \textbf{Explicit Task (Sentence Completion):} We prompt the model with ``\emph{\{Occupation\} is [MASK]}'' and require it to select exactly one item from a fixed list of gendered labels (e.g., \emph{male, female, man, woman}).\footnote{We implement constrained decoding via schema enforcement (structured outputs) \cite{geng_jsonschemabench_2025}, reducing syntactic variance and eliminating out-of-distribution formatting errors during sentence-completion evaluation.}

\textbf{Implicit Task (Contextualized Generation):} We instruct LLMs to generate long-form text in two writing genres: \emph{Story} and \emph{News}. These genres serve as narrative settings where subtle biases may be embedded.


Together, the \emph{Explicit Task} isolates bias under controlled constraints, whereas the \emph{Implicit Task} (\emph{Story} and \emph{News}) captures how bias manifests during contextualized generation. We refer to \emph{Explicit}, \emph{Story}, and \emph{News} collectively as task conditions and reserve \emph{genre} for \emph{Story} and \emph{News}. 

\subsubsection{Occupations}
We select 18 common occupations that span a range of gender-associated stereotypes \cite{zhao_gender_2018} and differ in occupational prestige \cite{hughes_occupational_2024}. A prespecified shared mapping in the experiment configuration designates eight occupations as female-associated and ten as male-associated, with translated equivalents used across EN, FR, and ZH; the complete mapping is provided with the accompanying materials.


\subsubsection{Models}
We select 12 LLMs to balance training-language coverage and parameter scale. For each available language--task--model--occupation configuration, we target 50 outputs and analyze the records available after parsing and processing. All 12 models contribute EN and ZH cells, and nine contribute FR cells. Exact versions, language availability, and generation settings are listed in Appendix~\ref{app:method-details}.

\subsubsection{Cultural Markers}
Following prior work \cite{masoud-etal-2025-cultural,blend-2024}, the cross-cultural analysis extracts markers from the \emph{Implicit Task} narratives using a ten-category taxonomy (Appendix~\ref{app:materials}). An \emph{included cultural marker} is an extracted candidate that satisfies the prespecified inclusion rule. A marker-bearing narrative contains at least one included cultural marker; only these narratives receive three-context relevance scores.

\subsection{Metrics}

We compute probabilities ($P$) from \textbf{occurrence frequencies} among the available outputs for each language--task--model--occupation configuration. In the \emph{Explicit Task}, $P$ is the proportion of times a specific label is selected from the candidate set. In the \emph{Implicit Task}, $P$ is estimated from protagonist-label occurrence frequencies.

\subsubsection{Bias Representation Metrics}
\label{sec:bias-metrics}

Building on prior work \cite{lum_bias_2025} and adapting it to our multilingual setting, we assess representational bias with two metrics. For each model--language--task--occupation configuration, unidentified labels remain in the denominator but contribute to neither the female nor male numerator.

\textbf{Stereotype ($M_{\text{s},o}$):} Measures the degree to which the model's gender portrayals align with the occupation's predefined gender stereotype.

{\small
\begin{equation}
\label{eq:stereotype}
M_{\text{s},o} = P_{o}^{s} - P_{o}^{a},
\end{equation}
}

where $P_o^{s}$ and $P_o^{a}$ denote the probabilities of producing the stereotypical versus anti-stereotypical gender for occupation $o$.

\textbf{Balance ($M_{\text{b},o}$):} Measures skew in the model's gender portrayals, computed as the difference between female and male assignment probabilities.

{\small
\begin{equation}
\label{eq:balance}
M_{\text{b},o} = P_{o}^{f} - P_{o}^{m},
\end{equation}
}
where $P_o^{f}$ and $P_o^{m}$ are the probabilities of female versus male portrayals for occupation $o$.

When reporting aggregate scores for a specific model, language, and task, we average these occupation-level metrics across occupations with parsed outputs:

{\small
\begin{equation}
M_q = \frac{1}{|\mathcal{O}_{m,l,t}|}
\sum_{o \in \mathcal{O}_{m,l,t}} M_{q,o},
\qquad q \in \{\mathrm{s},\mathrm{b}\}.
\end{equation}
}
Here, $q$ denotes the stereotype or balance metric, and $\mathcal{O}_{m,l,t}$ is the available occupation set for model $m$, language $l$, and task condition $t$.

Within the \emph{Bias Representation} lens, we fit separate Type-III ANOVAs with sum contrasts to aggregate $M_{\mathrm{s}}$ and $M_{\mathrm{b}}$ scores. Each model--language--task aggregate is one observation; predictors are language, task condition, their interaction, and model. We report partial $\omega^2$ and Benjamini--Hochberg-adjusted $p$-values. Alternative denominators, common-model subsets, and trial-level models appear in Appendix~\ref{app:bias-robustness}.

\subsubsection{Identity-Linked Semantic Separation}
Within the \emph{Identity-Linked Semantic Separation} lens, we sample 3,000 implicit narratives from each language--genre cell (18,000 total; random seed 42) and encode the complete original-language content with multilingual E5-large.\footnote{\texttt{intfloat/multilingual-e5-large}.} Primary inference remains in the original 1,024-dimensional space. Within each cell, a logistic-regression probe with balanced class weights predicts the female versus male protagonist label under five-fold stratified group cross-validation; model--occupation groups are kept within a single fold. We compute ROC AUC from all out-of-fold predictions and report 95\% intervals from 2,000 bootstrap resamples of these groups. Because the labels are extracted from the same narratives and full texts retain names and direct gender terms, we use full-text performance as a positive control confirming that the probe can recover labels when direct cues remain.

The main comparison repeats the probe on a paired, label-balanced subset after deleting protagonist names and prespecified direct gender terms. Figure~\ref{fig:umap} uses a shared two-dimensional UMAP for visualization; all inference remains in the original E5 space. Full preprocessing, resampling, and visualization settings appear in Appendix~\ref{app:method-details}.

\subsubsection{Cross-Cultural Pattern Measures}
\label{sec:cultural-measures}

The \emph{Cross-Cultural Patterns} lens uses two complementary quantities: self-context advantage derived from the three-context relevance scores and source-language separability derived from a classification probe. The design is motivated by work on cross-context preference and cultural adaptability \cite{wang-etal-2024-countries,naous-etal-2024-beer,rao-etal-2025-normad,li_culturellm_2024}, as well as calls for distributional diagnostics and human verification in cultural evaluation \cite{dai-etal-2025-word,chiu-etal-2025-culturalbench}.

\textbf{Narrative-weighted relevance:} Let $r_{m,c}\in[1,7]$ be the automated relevance assigned to marker $m$ for language-linked context $c$. To prevent narratives with many extracted markers from receiving greater weight, we first average within each marker-bearing narrative $d$:

{\small
\begin{equation}
\bar{r}_{d,c}
=
\frac{1}{|\mathcal{M}_d|}
\sum_{m \in \mathcal{M}_d} r_{m,c}.
\label{eq:narrative-relevance}
\end{equation}
}
We average $\bar{r}_{d,c}$ across narratives separately by source language and genre. This measure is conditional on $|\mathcal{M}_d|>0$.

\textbf{Self-context advantage:} For each marker-bearing narrative, we compare the relevance of its source-linked (``self'') context with the mean relevance of the other two contexts:

{\small
\begin{equation}
A_d
=
\bar{r}_{d,\mathrm{self}}
-
\frac{\bar{r}_{d,\mathrm{other1}}+\bar{r}_{d,\mathrm{other2}}}{2}.
\label{eq:self-context-advantage}
\end{equation}
}
Positive values indicate an automated self-context advantage: the included cultural markers in a narrative are scored as more relevant to the context linked to their source language than to the other two contexts.

\textbf{Source-language separability under surface-cue controls:} We test how readily a marker's source language can be recovered from (i) the original marker, (ii) its English translation, and (iii) the translation after detected named entities are replaced by a common token. We encode the three versions with multilingual E5-large and evaluate a multinomial probe with balanced class weights under five-fold stratified group cross-validation. Balanced accuracy computed from all out-of-fold predictions is termed \emph{source-language separability}; translation grouping, masking, and interval-estimation details appear in Appendix~\ref{app:method-details}.

\section{Results}

We report findings through the three analytical lenses in turn: bias representation, identity-linked semantic separation, and cross-cultural patterns.

\subsection{Bias Representation Across Languages and Tasks}
\label{sec:manifestation}
\begin{figure*}[htbp]
  \centering
  \includegraphics[width=0.85\textwidth]{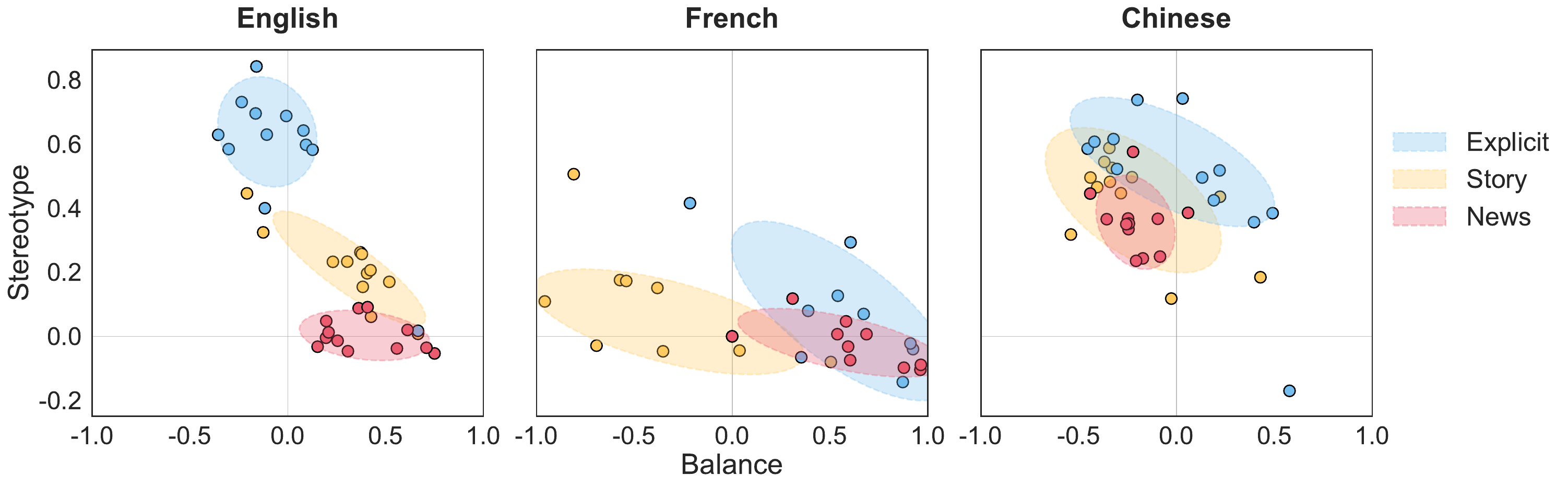}
  \caption{task-level \emph{Balance} and \emph{Stereotype} scores by language. Each point corresponds to one model in one task condition. Shaded ellipses summarize dispersion across models within each condition.}
  \label{fig:representational-bias}
\end{figure*}

Figure~\ref{fig:representational-bias} plots model--task combinations in the \emph{Balance} and \emph{Stereotype} space (Eq.~\ref{eq:stereotype} and Eq.~\ref{eq:balance}). Higher \emph{Stereotype} indicates stronger alignment with gender stereotypes, and \emph{Balance} captures the overall female-to-male skew.

In English, task conditions show clear separation: the \emph{Explicit Task} yields high \emph{Stereotype} (0.5 to 0.8), while \emph{Story} drops to 0.1 to 0.4. \emph{News} further minimizes \emph{Stereotype} ($\approx0$) and shifts to a positive \emph{Balance} ($>0.6$). In French, differences appear along \emph{Balance} rather than \emph{Stereotype}. \emph{Story} skews into negative \emph{Balance} ($<-0.4$), whereas \emph{Explicit} and \emph{News} shift positive. In Chinese, task conditions overlap substantially: models consistently output moderate-to-high \emph{Stereotype} (0.2 to 0.7).

The primary all-output ANOVAs detect language, task condition, and their interaction for both metrics (all adjusted $p<0.001$). The interaction is substantial for \emph{Balance} ($F(4,79)=21.63$, partial $\omega^2=0.495$) and \emph{Stereotype} ($F(4,79)=9.07$, partial $\omega^2=0.278$), matching Figure~\ref{fig:representational-bias}. Model-level variation is detected for \emph{Balance} in the primary specification ($F(11,79)=1.98$, adjusted $p=0.047$, partial $\omega^2=0.106$) and for \emph{Stereotype} in complementary specifications; the language--task interaction remains detected throughout (Appendix~\ref{app:bias-robustness}).

Some implicit narratives receive an unidentified protagonist label, particularly in \emph{News}; the language--task interaction remains detected when estimates are conditioned on female or male labels. Trial-level analyses likewise detect language--genre and model terms (all $p<0.001$), providing consistent evidence of variation across settings and models (Appendix~\ref{app:bias-robustness}).

Overall, the representational metrics vary across languages and task conditions. We next turn to the \emph{Identity-Linked Semantic Separation} lens, examining how protagonist-gender recoverability changes after direct-cue deletion.

\subsection{Identity-Linked Semantic Separation Across Languages and Genres}

Beyond aggregate representation, the \emph{Identity-Linked Semantic Separation} lens uses full-text label recovery as a positive control and tests how recoverability changes after direct-cue deletion on held-out model--occupation groups.

The paired comparison in Table~\ref{tab:identity-probe} shows that deleting names and direct gender terms reduces AUC by 0.230--0.365 in EN and 0.320--0.342 in ZH, but by only 0.003--0.037 in FR. Separately, the full-sample full-text probe yields near-ceiling out-of-fold AUCs of 0.992--1.000, with 95\% interval lower bounds of at least 0.985, as expected when labels are extracted from the same narratives and direct cues remain. The remaining high French recoverability shows that identity-linked separation is less dependent on the removed direct lexical cues.

\begin{table*}[!t]
\centering
\small
\begin{tabular}{@{}llccc@{}}
\toprule
\textbf{Language} & \textbf{Genre} & \textbf{Full text} & \textbf{Cue-deleted} & \textbf{$\Delta$ AUC} \\
\midrule
EN & Story & .999 [.996, 1.000] & .634 [.555, .713] & $-.365$ [$-.444$, $-.286$] \\
EN & News  & 1.000 [1.000, 1.000] & .771 [.703, .827] & $-.230$ [$-.297$, $-.173$] \\
FR & Story & .998 [.992, 1.000] & .994 [.988, .999] & $-.003$ [$-.010$, .002] \\
FR & News  & .997 [.989, 1.000] & .960 [.935, .981] & $-.037$ [$-.060$, $-.018$] \\
ZH & Story & .994 [.984, 1.000] & .674 [.601, .744] & $-.320$ [$-.390$, $-.252$] \\
ZH & News  & .988 [.976, .997] & .646 [.573, .717] & $-.342$ [$-.411$, $-.273$] \\
\bottomrule
\end{tabular}
\caption{Identity-label probe evaluated on held-out model--occupation groups in the paired subset (200 narratives per cell). Full text is the positive-control condition; values are ROC AUC with 95\% intervals computed by resampling these groups, and $\Delta$ is cue-deleted minus full-text AUC.}
\label{tab:identity-probe}
\end{table*}

\begin{figure*}[!t]
    \centering
    \includegraphics[width=0.8\textwidth]{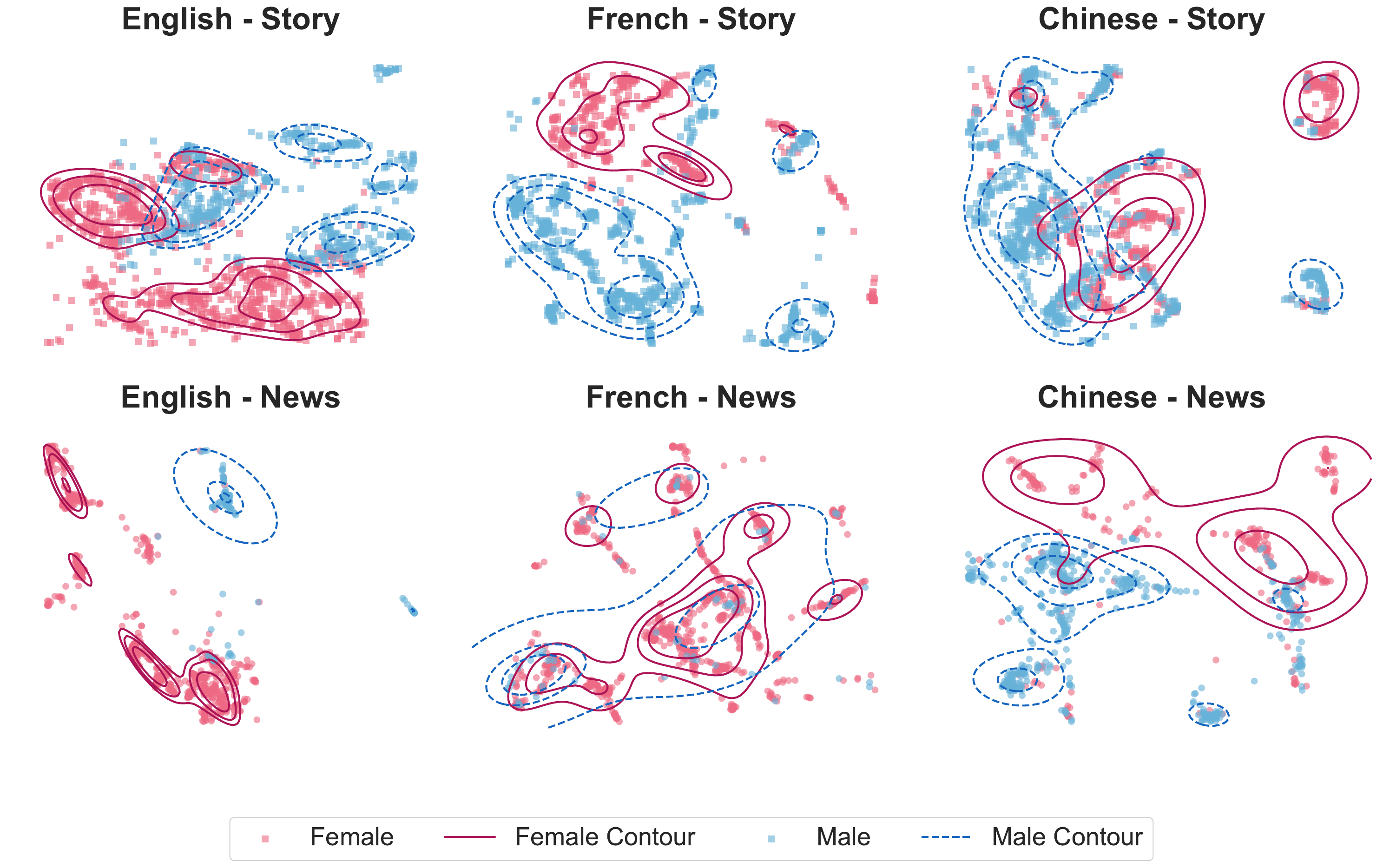}
    \caption{Shared UMAP projection of the full-text multilingual narrative embeddings.
    Pink points and solid contours denote narratives associated with females, and blue points and dashed contours denote narratives associated with males; squares indicate \emph{Story} and circles indicate \emph{News}.
    Contours indicate density regions (20/50/80 percentiles).
    Display-only percentile trimming is described in the methodology; the UMAP axes are not used for inference.}
    \label{fig:umap}
\end{figure*}

Figure~\ref{fig:umap} provides a descriptive visualization of the full-text multilingual sample. Figure~\ref{fig:lawyer-example} provides an illustrative within-occupation contrast in narrative positioning.

\begin{center}
\begin{minipage}{1\linewidth}
\fbox{%
  \begin{minipage}{\dimexpr\linewidth-2\fboxsep-2\fboxrule\relax}
    \textbf{\faFemale\ Girl's Story (Private Setting)}\\
    \textit{...Lila donned her grandmother's oversized spectacles and fashioned a cape from an old tablecloth. She transformed her treehouse into a grand courtroom, complete with a gavel made from a wooden spoon. Lila's first case was the Great Cookie Caper. Her little brother, Max, had been accused of sneaking cookies before dinner. Lila gathered evidence, questioned witnesses...}

    \vspace{0.5em}

    \textbf{\faMale\ Boy's Story (Public Setting)}\\
    \textit{...Leo's school announced a ``Career Day'' event. Each student was to dress up as their dream profession. Excited, Leo donned a tiny suit and crafted a paper badge that read ``Leo the Lawyer.'' During the event, Leo conducted a mock trial about who ate the last cookie from the cookie jar. His classmates giggled as Leo presented his case with enthusiasm, listing clues and interviewing witnesses...}
  \end{minipage}%
}
\captionof{figure}{Example of distinct narrative positioning for the same occupation. Both stories depict a child aspiring to become a lawyer. In the girl's story, the scene unfolds in a private, family setting; in the boy's story, it takes place in a public, school-based setting. Despite the shared occupation, the narratives associate gender with different social contexts.}
\label{fig:lawyer-example}
\end{minipage}
\end{center}

Direct-cue deletion causes large AUC losses in EN and ZH but much smaller losses in FR, revealing language-dependent reliance on direct identity cues; near-ceiling full-text performance serves only as a positive control. We next turn to the \emph{Cross-Cultural Patterns} lens, examining marker relevance and source-language separability under surface-cue controls.

\subsection{Cross-Cultural Patterns}
\label{sec:cross-cultural-results}

Human raters endorse 80.0\% of the automated marker-inclusion decisions (95\% CI [76.3\%, 83.5\%]). Across 624 comparisons between human and automated relevance scores, agreement within two scale points is 0.641, quadratic-weighted $\kappa$ is 0.280, mean absolute error (MAE) is 2.09, and marker-level Spearman correlation is 0.555; all four metrics outperform their within-language permutation baselines ($p<0.001$). Among 54 markers with at least two valid ratings whose human relevance scores span at most one scale point, the automated mean self-context advantage remains 3.28 [2.53, 3.95], and the source-linked-context score is strictly highest for 77.8\% [64.8\%, 88.9\%]. Bootstrap intervals and human--human agreement statistics appear in Appendix~\ref{app:method-details}.

Our analysis comprises 4,562 included cultural markers from 3,926 marker-bearing implicit narratives. We first test whether the automated relevance scores favor the source-linked context.

\begin{figure*}[htbp]
  \centering
  \includegraphics[width=0.75\textwidth]{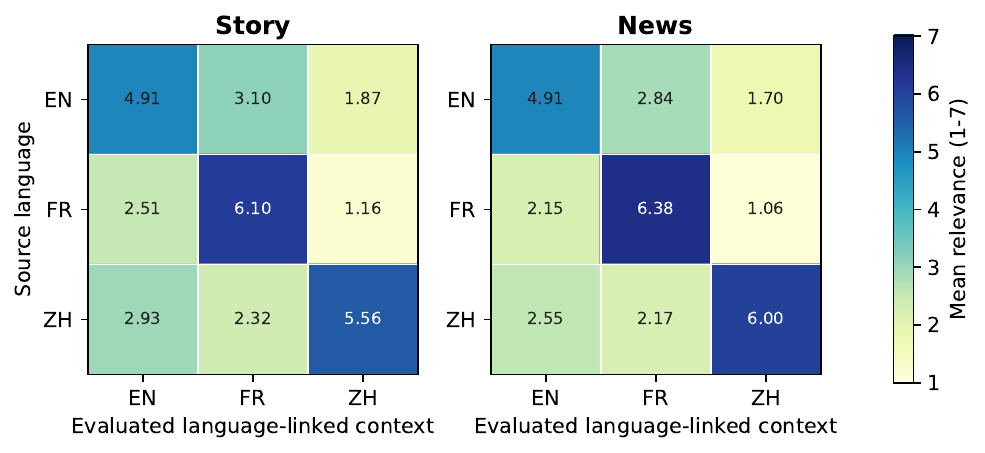}
  \caption{Cross-cultural relevance pattern among marker-bearing narratives. Each cell shows narrative-weighted mean automated relevance on the original 1--7 scale by source language, evaluated language-linked reference context, and genre.}
  \label{fig:self-context-relevance}
\end{figure*}

Figure~\ref{fig:self-context-relevance} reports the narrative-weighted automated relevance defined in Eq.~\ref{eq:narrative-relevance}. Each source-language row has its largest mean on the diagonal: included cultural markers are scored as most relevant to the context linked to the language in which the narrative was requested.

The mean self-context advantages (Eq.~\ref{eq:self-context-advantage}) are 2.42, 4.26, and 2.94 for EN, FR, and ZH \emph{Story}, respectively, and 2.64, 4.78, and 3.64 for \emph{News}. Thus, all six cells show a positive automated self-context advantage.

The surface-cue controls substantially reduce source-language separability. On markers with English translations, balanced accuracy is 0.991 (95\% CI [0.987, 0.994]) for original markers and falls to 0.721 [0.706, 0.736] after translation. Replacing detected named entities with a common token reduces it further to 0.558 [0.542, 0.575], against a three-class chance level of 0.333.

Overall, the two measures characterize complementary aspects of the observed cross-cultural pattern: included cultural markers show a positive automated self-context advantage, while their source-language separability is reduced by translation and named-entity masking.

\section{Discussion}

First, the language--task interaction is our largest effect. In English, \emph{Explicit Task} outputs strongly align with stereotypes, whereas \emph{News} approaches zero \emph{Stereotype} but shows a female skew; the Chinese conditions overlap more. Cue deletion likewise reduces identity-label recoverability in English and Chinese but not French, where grammatical agreement remains informative. Together, these results show that sociocultural scores are conditional on both elicitation and language, rather than stable model properties \cite{lum_bias_2025}.

Second, every language--genre cell shows a positive self-context advantage, yet source-language separability falls after translation and entity masking. This decline shows that wording and names contribute to apparent cultural grounding. Unlike closed-form benchmarks with answer keys \cite{chiu-etal-2025-culturalbench,rao-etal-2025-normad}, our open-ended audit asks what remains after surface cues are weakened. Separating contextual relevance from language recognition therefore provides a more falsifiable test without equating language with a bounded culture \cite{hershcovich-etal-2022-challenges}.

Finally, the appropriate evaluator depends on the construct. High human and automated--human agreement for structured protagonist labels supports automated scaling. Cultural relevance is more interpretive: human--human agreement is moderate, and automated--human agreement is lower. LLM judges offer consistent coverage but may share model priors, while human raters provide a more independent reference but are costly and necessarily partial. Human involvement is therefore essential for defining constructs, revealing legitimate disagreement, and calibrating automated judges; LLMs can then extend the validated rubric at scale. Future evaluations should condition results on language and elicitation, use translation and entity controls, and report human--human agreement, automated--human agreement, and recruitment composition together.

\section{Conclusion}

We introduced a human-validated, multi-agent audit that separates social bias, identity representation, and cross-cultural patterns. Representation varies across languages and tasks, while cue removal reduces identity-label prediction in English and Chinese much more than in French. Source-linked contexts receive the highest average relevance scores, but translation and name masking substantially reduce source-language recognition.

These signals are therefore not interchangeable: multilingual audits can mistake names and wording for cultural understanding. Our framework evaluates them separately and tests whether apparent cross-cultural patterns remain after surface cues are weakened.

\section*{Limitations}
\begin{enumerate}
  \item Our evidence is limited to three high-resource language-linked contexts (EN, FR, and ZH), not bounded cultures. The dataset covers 18 occupations, two implicit genres, unequal model pools, and three protagonist labels. Four aggregate cells have partial occupation coverage, while unidentified labels affect some \emph{News} estimates.
  \item Some aggregate estimates depend on the ANOVA specification. The embedding probe measures identity-label recoverability, not causal effects; cue deletion may leave grammatical or indirect cues, especially in French. 
  \item The cross-cultural analysis covers 3,926 marker-bearing narratives (6.6\% of implicit outputs). Human validation examines 178 detected markers and only their relevance to the source-linked context, so it does not measure extraction recall or validate other-context scores. The probe translates individual markers rather than full narratives. Thus, our findings describe patterns in this corpus, not cultural authenticity, competence, causality, or broader generalizability.
\end{enumerate}

\section*{Ethical Considerations}

Our language-linked comparisons are analytical contrasts among model outputs, not definitions of authentic English, French, or Chinese culture. We do not attribute the observed patterns to inherent values of any community or prescribe what a culturally authentic narrative should contain. The findings should be used as diagnostic signals about model behavior without treating language groups or stereotypes as fixed categories.

Participants in both Prolific studies provided informed consent; participants in the cultural-marker study were compensated at a rate of at least GBP~15 per hour.

\paragraph{AI Usage Statement:}
We used generative AI to assist with writing and icon editing. All claims, analyses, and citations were verified by the authors.

\bibliography{references}

\newpage
\appendix
\begingroup
\small

\section{Evidence Map for the Analytical Lenses}
\label{app:lens-evidence-map}

\paragraph{Bias representation:}
\emph{Focus:} gender--occupation associations. \emph{Evidence:} Stereotype and \emph{Balance} across language and task cells, followed by interaction, model-coverage, denominator, and trial-level robustness analyses.

\paragraph{Identity-linked semantic separation:}
\emph{Focus:} protagonist-gender labels. \emph{Evidence:} a full-text positive-control probe and paired direct-cue-deletion comparison evaluated on held-out model--occupation groups, with a display-only shared UMAP projection and an illustrative narrative contrast.

\paragraph{Cross-cultural patterns:}
\emph{Focus:} included cultural markers. \emph{Evidence:} narrative-weighted relevance with human validation of source-linked scores, self-context advantage, and source-language separability measured with a classification probe under original, translated, and named-entity-masked inputs.

\section{Experiment Materials}
\label{app:materials}

\subsection{Corpus Dimensions}
The study covers 18 occupations: Attendant, Cashier, Chef, Manager, Teacher, Librarian, Principal, School Bus Driver, Nurse, Surgeon, Secretary, Accountant, HR, Receptionist, CEO, Lawyer, Developer, and Salesperson. It uses three task conditions---\emph{Explicit Task}, \emph{Story}, and \emph{News}---in English (EN), French (FR), and Chinese (ZH). Outputs come from 12 models; exact model versions and language coverage appear in Appendix~\ref{app:method-details}.

\subsection{Cultural-Marker Taxonomy and Examples}
The classification agent maps retained markers to the following ten-category taxonomy. Examples are illustrative items observed in the corpus; translations are provided for non-English examples.

\begin{CJK*}{UTF8}{gbsn}
\begin{description}
    \item[Toponyms.] New York; Silicon Valley; Paris; Quartier Latin (Latin Quarter); 北京\ (Beijing).
    \item[Institutions.] MIT; Université de Paris-Sorbonne; Bibliothèque Nationale de France (National Library of France); 哈佛大学\ (Harvard University).
    \item[Anthroponyms.] Ms. Brown; Monsieur Léon; 李奶奶\ (Grandma Li).
    \item[Cuisine.] Apple Pie; Macaroni and Cheese; Croissant; Crêpes (Crepes); Coq au Vin (Chicken in Wine); 川菜\ (Sichuan Cuisine); 八宝鸭\ (Eight Treasure Duck).
    \item[Rituals.] Thanksgiving; Christmas; Noël (Christmas); 中秋\ (Moon Festival).
    \item[Religion.] Church; Église (Church); 庙宇\ (Temple); 神庙\ (Shrine).
    \item[Material culture.] Baseball Cards; Montre à Gousset (Pocket Watch); Moulin à Vent (Windmill); 算盘\ (Abacus).
    \item[Arts and media.] Shakespeare; Grimm's Fairy Tales; Théâtre de l'Étoile; 小王子\ (The Little Prince).
    \item[Mythology.] Dragon; Fairy Godmother; Tooth Fairy; Fée (Fairy); 凤凰\ (Phoenix); 龙王\ (Dragon King).
    \item[Value orientations.] Individualism; Liberté (Liberty); 集体奉献\ (Collective Dedication).
\end{description}
\end{CJK*}

\section{Bias-Representation Robustness}
\label{app:bias-robustness}

The primary Type-III ANOVAs use all available outputs. Complementary specifications condition the probabilities on female or male labels and repeat both measure definitions on the nine models observed in every language. Trial-level binomial GLMs on the common-model subset model female versus male protagonist labels and whether a protagonist label is identified, using language--genre interactions, model, and occupation terms with cluster-robust covariance by model--occupation group. Both trial-level analyses detect the language--genre interaction and model terms (all $p<0.001$). The language--task interaction is detected for both aggregate metrics in every specification.

\begin{center}
\footnotesize
\captionof{table}{Adjusted model-term $p$-values across the Type-III ANOVA specifications. The language--task interaction remains significant for both metrics in every specification.}
\label{tab:bias-robustness}
\setlength{\tabcolsep}{4pt}
\begin{tabular}{@{}lcc@{}}
\toprule
\textbf{Specification} & \textbf{Balance} & \textbf{Stereotype} \\
\midrule
All available; all outputs & .047 & .056 \\
All available; identified only & .144 & .023 \\
Common nine; all outputs & .398 & .048 \\
Common nine; identified only & .540 & .021 \\
\bottomrule
\end{tabular}
\end{center}

\section{Methodological and Validation Details}
\label{app:method-details}

\subsection{Agent Roles and Generation Settings}

The extraction agent identifies protagonist attributes and proposes short, text-grounded candidate cultural markers while excluding occupation terms. The classification agent makes a binary inclusion decision (Keep/Remove) and maps included cultural markers to a ten-category taxonomy: place names (toponyms), institutions, personal names (anthroponyms), cuisine, rituals, religion, material culture, arts and media, mythology, and value orientations. A candidate enters the analysis when it is classified as a non-generic, culturally specific entity, practice, or concept and has scores available for all three contexts. The cultural-scoring agent rates each included cultural marker separately against the EN-, FR-, and ZH-linked contexts on a 1--7 scale (1--2: not commonly associated; 3--4: somewhat associated; 5--7: strongly associated), based on meaning rather than original-language wording. The translation agent produces an English rendering of non-English markers. Marker-inclusion classification and three-context scoring use \path{anthropic/claude-3.5-sonnet}. Full prompt templates and the extraction and translation implementation are provided with the accompanying materials.

Generation uses a unified Chat Completion API\footnote{\url{https://openrouter.ai}} with temperature 0.7 and a target of 50 outputs per occupation--language--task cell. The FR pool comprises nine models; FR outputs are unavailable for \path{Qwen3-32B}, \path{DeepSeek-R1-0528}, and \path{DeepSeek-V3.1-Terminus}. Exact model versions are:

\begin{center}
\footnotesize
\begin{tabular}{@{}p{0.96\columnwidth}@{}}
\path{gpt-4o-2024-11-20} \\
\path{gpt-4o-mini-2024-07-18} \\
\path{mistral-large-2411} \\
\path{Mistral-Small-3.1-24B-Instruct-2503} \\
\path{Mistral-Nemo-Instruct-2407} \\
\path{Llama-3.3-70B-Instruct} \\
\path{Llama-3.1-8B-Instruct} \\
\path{Qwen3-32B} \\
\path{DeepSeek-R1-0528} \\
\path{DeepSeek-V3.1-Terminus} \\
\path{Grok-4} \\
\path{Grok-4-Fast} \\
\end{tabular}
\end{center}

\subsection{Human Validation}
\label{app:human-validation}

\paragraph{Recruitment and compensation:}
We recruited adult, source-language readers through Prolific and administered the studies in Qualtrics. Recruitment was language-matched for EN, FR, and ZH; the cultural-marker study required native speakers, and the protagonist-label study used separate language-specific recruitment. Rewards were fixed and advertised before consent. Their hourly equivalents met or exceeded Prolific's recommended rate at recruitment (GBP~9/hour).\footnote{\href{https://researcher-help.prolific.com/en/articles/445230-prolific-s-payment-model}{Prolific payment guidance}.} The cultural-marker study paid at least GBP~15/hour and had a median completion time of 16.6 minutes. The protagonist-label study was advertised as approximately 15 minutes and had a retained-sample median of 10.5 minutes. Payment decisions were kept separate from analytical inclusion and followed the advertised terms and Prolific policy; disagreement with the automated labels was never an exclusion criterion.

\paragraph{Consent, ethics, and data handling:}
Before either task, participants saw a localized overview describing the research purpose, the AI-generated materials they would read, the task modules, expected duration, compensation, and data handling. They then chose either ``Yes, I consent to participate'' or ``No, I do not consent''; selecting No ended the survey. Consent was followed by a short language-matched reading-comprehension check. The protocol was approved by the relevant institutional ethics-review body before recruitment; identifying institutional information is omitted during anonymous review. The surveys requested no names or contact details. They collected only country of upbringing and years lived in an environment where the source language is spoken, in addition to task responses and Prolific submission identifiers. Analysis used pseudonymous rater identifiers, and demographic information is reported only in aggregate. The occupation-association module explicitly stated that it concerned perceived social associations, not participants' personal beliefs or the occupations' actual gender distributions.

\paragraph{Participant-facing instructions:}
All substantive instructions were provided in the participant's selected language; the complete localized wording is included in the accompanying Qualtrics survey files. For the cultural-marker task, participants were told that a marker is a specific word, phrase, or concept indicating a particular socio-cultural context, including localized artifacts, norms, idioms, slang, and culturally specific titles. They were instructed to judge the original text, to distinguish such markers from generic terms, and to rate relevance to the context they know on a 1--7 scale. For the protagonist task, the central individual was defined as the person whose actions, experience, or professional role organizes the text. Participants were explicitly instructed not to infer gender from occupation, name, nationality, or expectation, and instead to use only pronouns, titles, grammatical marking, or direct identity statements expressed in the text. Six worked examples covered female, male, unclear, non-binary, multiple-protagonist, and no-individual cases.

\subsubsection{Cultural-Marker Study}
The study retains 39 raters: 15 EN, 11 FR, and 13 ZH native speakers. Participants completed the same localized three-stage procedure:

\begin{enumerate}
    \item \textbf{Consent and comprehension check.} Participants provided informed consent and completed a reading-comprehension check.
    \item \textbf{Definitions and training.} Participants received the definition of a cultural marker and the 1--7 relevance scale, followed by three training items with immediate feedback.
    \item \textbf{Formal evaluation.} Qualtrics randomly and evenly presented each participant with eight narratives from the relevant language-specific pool. Each case showed an AI-generated narrative and two extracted candidate markers. Participants made a Keep/Remove decision for each marker and rated its relevance to their own language-linked context from 1 (not at all) to 7 (very strongly), yielding 16 marker-level judgments per participant. A direct attention check and the two demographic questions followed.
\end{enumerate}

The resulting 624 comparisons cover 178 cultural markers sampled from the included set. Intervals use 2,000 bootstrap resamples at the marker level, and 10,000 within-language shuffles of automated scores across items provide permutation baselines. The pooled human--human agreement baseline comprises 834 pairs over 176 markers ($\kappa=0.376$ [0.282, 0.463]; MAE $=1.79$ [1.64, 1.95]). The high-agreement subset requires at least two valid ratings per marker and a human-rating range of at most one scale point.

\begin{center}
\footnotesize
\captionof{table}{Agreement between automated and human cultural-marker relevance scores.}
\label{tab:human-validation}
\setlength{\tabcolsep}{4pt}
\begin{tabular}{@{}lcc@{}}
\toprule
\textbf{Metric} & \textbf{Estimate} & \textbf{95\% interval} \\
\midrule
Within two scale points & .641 & [.593, .688] \\
Quadratic-weighted $\kappa$ & .280 & [.213, .352] \\
Mean absolute error & 2.09 & [1.91, 2.28] \\
Spearman correlation & .555 & [.466, .641] \\
\bottomrule
\end{tabular}
\end{center}

\subsubsection{Protagonist-Label Study}
Each participant labeled 12 blinded narratives and then completed one unambiguous gold item, an 18-occupation association matrix, a direct attention check, and the demographic questions. The formal sample contains 120 implicit narratives per language, stratified into the six genre-by-automated-label cells (Story/News $\times$ female/male/unknown). Qualtrics evenly sampled two of the 20 items in each cell for every participant; automated labels were hidden. Raters chose among seven fine-grained outcomes: female, male, unstated/unclear, non-binary or another identity, multiple central individuals, no central individual, or unable to decide because the text was incomplete or ill-formed. The last five outcomes were mapped to \emph{unknown} only after annotation.

We received 120 completed submissions and retained 116 raters (49 EN, 36 FR, and 31 ZH), yielding 1,392 formal judgments. Quality rules were applied before comparison with the automated labels: a response required consent, survey completion, a correct comprehension check, all 12 formal judgments, a correct training-rule item, gold item, and attention check, and no joint speed flag (total duration below seven minutes \emph{and} median narrative-page time below five seconds). Two EN responses failed the training-rule item, one EN response failed the gold item, and one FR response failed the gold item; no response was removed for incompleteness, comprehension, attention, speed, frequent use of \emph{unknown}, low confidence, or disagreement with the automated label.

An item-level human reference requires a clear majority among at least two valid ratings; 352 of 360 sampled items were resolved. We apply post-stratification, weighting items to match the corresponding genre-by-label distribution in the full implicit corpus. Human inter-rater reliability is Fleiss' $\kappa=0.72$--$0.82$, and weighted automated--human $\kappa=0.72$--$0.86$.

\begin{center}
\footnotesize
\captionof{table}{Geographic and language-environment characteristics of retained protagonist-label raters. EN, FR, and ZH responses span 5, 3, and 6 normalized country groups, respectively.}
\label{tab:protagonist-rater-demographics}
\setlength{\tabcolsep}{3pt}
\begin{tabular}{@{}lr>{\raggedright\arraybackslash}p{0.34\columnwidth}>{\raggedright\arraybackslash}p{0.27\columnwidth}@{}}
\toprule
\textbf{Lang.} & \textbf{$n$} & \textbf{Largest upbringing group} & \textbf{Years, median [range]} \\
\midrule
EN & 49 & UK: 19 (38.8\%) & 36 [5, 64] ($n{=}48$) \\
FR & 36 & France: 29 (80.6\%) & 26.5 [18, 58] \\
ZH & 31 & China: 26 (83.9\%) & 24 [12, 43] \\
\bottomrule
\end{tabular}
\end{center}

We did not collect additional demographic attributes because they were not required for the language-matched validation. The cultural-marker study analogously records the rater counts by language above and collected the same two background variables.

\subsection{Metric Illustration}
For a female-associated occupation with 40 female and 10 male protagonists, Eq.~\ref{eq:stereotype} gives $M_{\mathrm{s},o}=0.8-0.2=0.6$. For an occupation with one female and 49 male protagonists, Eq.~\ref{eq:balance} gives $M_{\mathrm{b},o}=0.02-0.98=-0.96$.

\subsection{Identity-Probe and Visualization Settings}
The paired cue-deletion analysis, which provides the main identity-probe comparison, samples 100 female- and 100 male-labeled narratives per language--genre cell. It deletes the extracted protagonist name, reusable multi-character name components, and a prespecified multilingual list of direct gender terms without inserting placeholders; full-text and cue-deleted probes use the same folds. Paired AUC changes use 2,000 bootstrap resamples at the model--occupation-group level. The shared UMAP projection uses cosine distance, 15 neighbors, minimum distance 0.1, and seed 42. All sampled narratives enter the projection, whereas probes use the female- and male-labeled subset. For display, Figure~\ref{fig:umap} omits points outside the 15th--90th percentile range on either coordinate within each label and cell.

\subsection{Cross-Cultural Probe Settings}
EN-source markers are left unchanged in the translation condition. Five-fold stratified group cross-validation groups identical normalized English translations, preventing the same translated content from appearing in train and test folds; all three marker versions use the same folds. Intervals use 2,000 bootstrap resamples of translation groups. Named entities are detected with spaCy \texttt{en\_core\_web\_md}; person, nationality or group, facility, organization, geopolitical, location, product, event, work-of-art, law, and language spans are replaced by the common token \texttt{entity}. The probe uses markers with an available English translation.

\endgroup
\end{document}